\pdfoutput=1

\documentclass[11pt]{article}

\usepackage[preprint]{acl}

\usepackage{times}
\usepackage{latexsym}

\usepackage[T1]{fontenc}

\usepackage[utf8]{inputenc}

\usepackage{microtype}

\usepackage{inconsolata}
\usepackage{amsmath} 
\usepackage{tabularx}
\usepackage{arydshln}
\usepackage{graphicx}

\usepackage{tikz-dependency}
\usepackage[cjk]{kotex}

\title{K-UD: Revising Korean Universal Dependencies Guidelines}

\author{
Kyuwon Kim$^{1}$~~
Yige Chen$^{2}$~~
Eunkyul Leah Jo$^{3,4}$~~ \large{\textbf{KyungTae Lim}$^{5}$}~~
\large{\textbf{Jungyeul Park}$^{3}$}~~
\large{\textbf{Chulwoo Park}$^{6}$}\\
$^{1}$Seoul National University, South Korea~~
$^{2}$The Chinese University of Hong Kong, Hong Kong\\
$^{3}$The University of British Columbia, Canada~~
$^{4}$Sorbonne Université, France\\
$^{5}$SeoulTech, South Korea~~
$^{6}$Anyang University, South Korea \\
{\tt guwon0406@snu.ac.kr} ~~~
{\tt yigechen@link.cuhk.edu.hk}~~~ 
{\tt eunkyul@student.ubc.ca}\\
{\tt ktlim@seoultech.ac.kr}~~~
{\tt jungyeul@mail.ubc.ca}~~~
{\tt cwpa@anyang.ac.kr}
}

\begin{document}
\maketitle
\begin{abstract}
Critique has surfaced concerning the existing linguistic annotation framework for Korean Universal Dependencies (UDs), particularly in relation to syntactic relationships.
In this paper, our primary objective is to refine the definition of syntactic dependency of UDs within the context of analyzing the Korean language. 
Our aim is not only to achieve a consensus within UDs but also to garner agreement beyond the UD framework for analyzing Korean sentences using dependency structure, by establishing a linguistic consensus model.
\end{abstract}

\section{Introduction} \label{intro}
{Universal Dependencies (UDs) \citep{nivre-EtAl:2016:LREC,nivre-EtAl:2020:LREC} adhere to a uniform framework for maintaining consistent grammar annotation across various languages.
Currently, UDs offer 245 treebanks in 141 languages (Version 2.12).\footnote{\url{https://universaldependencies.org}} 
One of the benefits of UDs is the possibility to construct a seamless multilingual system without the need for additional efforts.
Korean UDs have been proposed firstly under the name of Google's homogeneous syntactic dependency annotation (GSD) \citep{mcdonald-etal-2013-universal}, and Kaist \citep{choi-EtAl:1994}. 
The latter was originally created as a constituency treebank, has been employed for the purpose of Korean constituency parsing during  Statistical Parsing of Morphologically Rich Languages (SPMRL) \citep{seddah-EtAl:2013:SPMRL,seddah-kubler-tsarfaty:2014}, and was subsequently converted into dependency structures as an integral part of UDs. 
Other Korean UD efforts, such as \citet{chun-EtAl:2018:LREC}, incorporate the Penn Korean treebank \citep{han-EtAl:2002}, which has been converted from constituency-based to dependency-based.}

\begin{figure}[!ht]
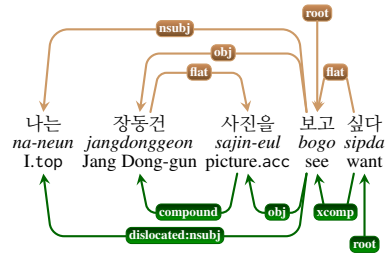

\centering
{
\begin{dependency}
{
{\scriptsize
   \begin{deptext}
나는\& 장동건\& 사진을\& 보고\& 싶다    \\
\textit{na-neun}\& \textit{jangdonggeon}\& \textit{sajin-eul} \& \textit{bogo} \& \textit{sipda} \\
I.\texttt{top} \& Jang Dong-gun \& picture.\texttt{acc} \& see \& want \\
   \end{deptext}
   \depedge[theme=copper,edge unit distance=3.3ex]{4}{1}{nsubj}
   \depedge[theme=copper,edge unit distance=3.5ex]{4}{2}{obj}
   \depedge[theme=copper]{2}{3}{flat}
   \deproot[theme=copper,edge unit distance=3.3ex]{4}{root}
   \depedge[theme=copper]{5}{4}{flat}
   \depedge[theme=grassy,edge below,edge unit distance=2.5ex]{4}{1}{dislocated:nsubj}
   \depedge[theme=grassy,edge below]{4}{3}{obj}
   \depedge[theme=grassy,edge below]{3}{2}{compound}
   \deproot[theme=grassy,edge below,edge unit distance=2.5ex]{5}{root}
   \depedge[theme=grassy,edge below]{5}{4}{xcomp}   
}
}
\end{dependency}
}
\caption{Example of the current annotation in \texttt{Korean\_GSD}}
\label{np-head-problem}
\end{figure}

However, criticism has arisen regarding the current linguistic annotation scheme for Korean UDs \citep{noh-etal-2018-enhancing}, especially syntactic relations \citep{deMarneffe-EtAl:2014:LREC}, for example, a noun phrase head (\textit{jangdong-geon} \textit{sajin-eul}) and an auxiliary verb construction (\textit{bogo} \textit{sipda}) as described in Figure~\ref{np-head-problem}.\footnote{By convention, we introduce `top' to refer to the current annotation and `bottom' to refer to the revised annotation as conventions in this paper when comparing them. All sentence examples up to Section~\ref{new-guidlines} are taken from the `ko\_gsd-ud' dataset (\texttt{ud-treebanks-v2.12}).}
In the former case, \textit{sajin-eul} (`picture.\texttt{acc}') is the central noun that represents the main object or concept in the phrase. The current Korean UDs always annotate the first noun as the head, regardless of their grammatical categories, whether they are proper nouns or common nouns.
Additionally, auxiliary verbs in Korean should be distinguished based on whether they are used in catenative constructions or to indicate tense. Currently, in Korean UDs, the first verb is always annotated as the head, irrespective of its syntactic and semantic properties. 
In the latter case, \textit{sipda} (`want') is a verb used in catenative constructions and serves as the lexical head of the entire sentence. 

The current Korean GSD dataset also exhibits morphological annotation errors, for example, with an average of 2.46 words per sentence for language-specific XPOS and 2.35 words per sentence for Universal POS, as reported in \citet{jo-etal-2023-k}.
In this paper, we focus on revising the definition of syntactic dependency with an attempt to establish the linguistic consensus for Korean language analysis not just within UDs, but also outside of UDs.

\section{Previous Work}

\subsection{{Korean dependency parsing datasets}}

Several resources of dependency parsing data for Korean have been made available in the past, including the GSD treebank \citep{mcdonald-etal-2013-universal}, the Kaist treebank \citep{choi-EtAl:1994,chun-EtAl:2018:LREC}, the Penn Korean universal dependency treebank\footnote{\url{https://catalog.ldc.upenn.edu/LDC2023T05}} \citep{han-EtAl:2002}, and the Korean Language Understanding Evaluation (KLUE) benchmark dependency parsing dataset \citep{park-etal-2021-klue}. Table \ref{tab:kor-dep-data} illustrates the size of each of the available datasets in terms of the number of sentences. 

\begin{table}[!ht]
\centering
\scriptsize
\begin{tabular}{c | ccc |c }\hline
        {Source} & {Train} & {Dev} & {Test} & {Total}\\\hline
        GSD & 4400 & 950 & 989 & 6339\\
        Kaist & 23,010 & 2066 & 2287 & 27,363\\
        Penn & --&--&--&5010\\
        KLUE & 10,000 & 2000 & 2500& 14,500\\\hline
    \end{tabular}
    \caption{Statistics of the available Korean dependency parsing datasets, in terms of the number of sentences.}
    \label{tab:kor-dep-data}
\end{table}

Dependency parsing data, derived from the Sejong constituency treebank, is available. 
Although various versions of these datasets have been generated by different studies, we do not enumerate them all in this paper to avoid redundancy. The GSD, Kaist, and Penn dependency treebanks are primarily in a UD-style `harmonized' format, while KLUE follows the Sejong-style, which includes `constituency-inherent' information in the dependency treebank.

\subsection{Representation strategies for Korean dependency parsing}
There have been previous studies trying to {address} the word-level representation issues of Korean \citep{choi-palmer:2011:SPMRL,park-EtAl:2013:IWPT,kanayama-EtAl:2014:LT4CloseLang}. 
CoNLL 2017-2018 shared tasks include Korean where every participant engages in Korean dependency parsing as a component of their multilingual parsing tasks \citep{zeman-etal-2017-conll,zeman-etal-2018-conll} {within the context of Universal Dependencies.} 
However, Korean is an agglutinative language {whose syntax and semantics} heavily {rely} on morphemes. The {fact that its natural segmentation does not correctly reflect either the word boundaries or the morpheme boundaries of Korean texts indicates that word-level representations are not ideal approaches. 
To address this issue, \citet{chen-etal-2022-yet} introduced an annotation scheme that breaks down Korean texts to the morpheme level. By applying this morpheme-based format to Korean dependency parsing, they achieved superior performance compared to word-based parsing models.

\section{Revised Korean UD Guidelines} \label{new-guidlines}
We try to focus on syntactic relations if there exists a disparity between the current Korean UDs and the suggested annotation scheme.

\subsection{Core arguments: nominals}
Nominal core arguments, such as  \texttt{nsubj} (nominal subject), \texttt{obj} (object) and \texttt{iobj} (indirect object), represent the syntactic roles within a clause. \texttt{nsubj:pass} (passive nominal subject) is also included in this category. 
These nominals are typically headed by a noun. In the case of a noun compound, the head is typically the stem that determines the semantic category, usually the last noun in Korean.

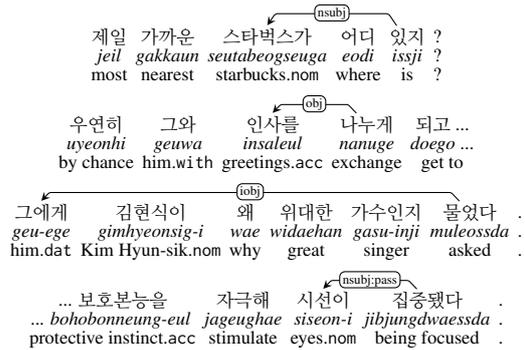
\begin{center}
\begin{dependency}
{
{\scriptsize
   \begin{deptext}
제일\& 가까운\& 스타벅스가\& 어디\& 있지\&?\\
\textit{jeil}\& \textit{gakkaun}\& \textit{seutabeogseuga}\& \textit{eodi}\& \textit{issji}\&? \\
most\& nearest\& starbucks.\texttt{nom}\& where\& is\& ? \\
   \end{deptext}
   \depedge[edge unit distance=.7ex]{5}{3}{nsubj}
}
}
\end{dependency}    
\begin{dependency}
{
{\scriptsize
   \begin{deptext}
우연히\& 그와\& 인사를\& 나누게\& 되고 ...\\ 
\textit{uyeonhi}\& \textit{geuwa}\& \textit{insaleul}\& \textit{nanuge}\& \textit{doego} ...\\
by chance\& him.\texttt{with}\& greetings.\texttt{acc}\& exchange\& get to \\
   \end{deptext}
   \depedge[edge unit distance=1ex]{4}{3}{obj}
}
}
\end{dependency}

\begin{dependency}
{
{\scriptsize
   \begin{deptext}
그에게\&   김현식이\&       왜\&   위대한\& 가수인지\& 물었다\& . \\
\textit{geu-ege}\& \textit{gimhyeonsig-i}\& \textit{wae}\& \textit{widaehan}\& \textit{gasu-inji} \& \textit{muleossda} \& . \\
him.\texttt{dat} \& Kim Hyun-sik.\texttt{nom} \& why \& great \& singer \& asked \& .\\
   \end{deptext}
   \depedge[edge unit distance=.2ex]{6}{1}{iobj}
}
}
\end{dependency}

\begin{dependency}
{
{\scriptsize
   \begin{deptext}
... 보호본능을 \& 자극해 \& 시선이 \&   집중됐다 \& . \\
... \textit{bohobonneung-eul } \& \textit{jageughae} \& \textit{siseon-i} \& \textit{jibjungdwaessda} \& . \\
protective instinct.\texttt{acc} \& stimulate \& eyes.\texttt{nom} \& being focused \& .\\
   \end{deptext}
   \depedge[edge unit distance=1ex]{4}{3}{nsubj:pass}
}
}
\end{dependency}
\end{center}

As we discussed in $\S$\ref{intro}, in the current Korean UDs annotation, the first noun in a noun phrase, especially in a compound noun, is typically annotated as the head and treated as the argument of the predicate. However, the revised guidelines propose annotating the last noun as the head. Even in the case of multiword proper names, especially in many languages with no clear internal syntactic structure, where the first proper noun is generally considered as the head, we annotate a last proper noun as the head in Korean. 
This choice is made because the last proper noun would receive the case from the predicate.


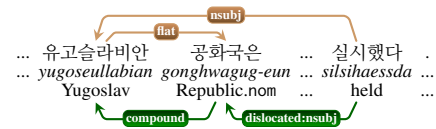
\begin{center}
\begin{dependency}
{
{\scriptsize
   \begin{deptext}
...\&  유고슬라비안\&  공화국은\&  ... \&  실시했다 \& . \\
...\& \textit{yugoseullabian} \& \textit{gonghwagug-eun} \& ... \& \textit{silsihaessda} \& ... \\
\& Yugoslav \& Republic.\texttt{nom} \& ... \& held \& ...\\
   \end{deptext}
   \depedge[theme=copper,edge unit distance=1ex]{5}{2}{nsubj}   
   \depedge[theme=copper,edge unit distance=1ex]{2}{3}{flat}   
   \depedge[theme=grassy,edge below,edge unit distance=1ex]{5}{3}{dislocated:nsubj}
   \depedge[theme=grassy,edge below,edge unit distance=2ex]{3}{2}{compound}
}
}
\end{dependency}
\end{center}

\subsection{Core arguments: clauses}

Clausal core arguments, such as \texttt{csubj} (clausal subject), 
\texttt{ccomp} (clausal complement) and  
\texttt{xcomp} (open clausal complement), 
represent the syntactic role where the argument is itself a clause. For example, a clausal subject is a clausal syntactic subject of a clause. 
 
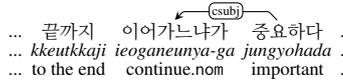
\begin{center}
\begin{dependency}
{
{\scriptsize
   \begin{deptext}
...\& 끝까지\& 이어가느냐가 \&중요하다 \&.\\
...\& \textit{kkeutkkaji} \& \textit{ieoganeunya-ga} \& \textit{jungyohada} \&  . \\
...\& to the end \& continue.\texttt{nom} \& important \& .\\
   \end{deptext}
   \depedge[edge unit distance=1ex]{4}{3}{csubj}
}
}
\end{dependency}
\end{center}

A clausal complement of a predicate is a dependent clause which is a core argument where we distinguish with \texttt{xcomp} used in catenative constructions. 
The clausal complement (\texttt{ccomp}) is a dependent clause that functions as a core argument, and we use the open clausal complement (\texttt{xcomp}) to distinguish it in catenative constructions. 

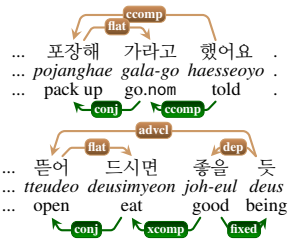
\begin{center}
\begin{dependency}
{
{\scriptsize
   \begin{deptext}
...\& 포장해\& 가라고 \&했어요 \&.\\
...\& \textit{pojanghae} \& \textit{gala-go} \& \textit{haesseoyo} \&  . \\
...\& pack up \& go.\texttt{nom} \& told \& .\\
   \end{deptext}
   \depedge[theme=copper,edge unit distance=1.5ex]{4}{2}{ccomp}
   \depedge[theme=copper,edge unit distance=1.5ex]{2}{3}{flat}
   \depedge[theme=grassy,edge below,edge unit distance=1ex]{4}{3}{ccomp}
   \depedge[theme=grassy,edge below,edge unit distance=1ex]{3}{2}{conj}
}
}
\end{dependency}

\begin{dependency}
{
{\scriptsize
   \begin{deptext}
...\& 뜯어\& 드시면 \&좋을 \&듯\\
...\& \textit{tteudeo} \& \textit{deusimyeon} \& \textit{joh-eul} \&  \textit{deus} \\
...\& open \& eat \& good \& being\\
   \end{deptext}
\depedge[theme=copper,edge unit distance=1ex]{5}{2}{advcl}
\depedge[theme=copper,edge unit distance=1ex]{2}{3}{flat}   
\depedge[theme=copper,edge unit distance=1ex]{5}{4}{dep}   
\depedge[theme=grassy,edge below,edge unit distance=1.5ex]{4}{3}{xcomp}
\depedge[theme=grassy,edge below,edge unit distance=1.5ex]{3}{2}{conj}
\depedge[theme=grassy,edge below,edge unit distance=1.5ex]{4}{5}{fixed}
}
}
\end{dependency}
\end{center}

While the current Korean UDs do not include \texttt{csubj:pass} (clausal passive subject), verbs that could have \texttt{nsubj:pass} can also take a \texttt{csubj:pass} relation with a clausal argument.

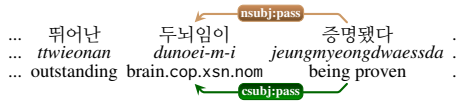
\begin{center}
\begin{dependency}
{
{\scriptsize
   \begin{deptext}
...\& 뛰어난\& 두뇌임이 \&증명됐다 \&.\\
...\& \textit{ttwieonan} \& \textit{dunoei-m-i} \& \textit{jeungmyeongdwaessda} \&  . \\
...\& outstanding \& brain.\texttt{cop}.\texttt{xsn}.\texttt{nom} \& being proven \& .\\
   \end{deptext}
    \depedge[theme=copper,edge unit distance=1ex]{4}{3}{nsubj:pass}   
   \depedge[theme=grassy,edge below,edge unit distance=1ex]{4}{3}{csubj:pass}   
}
}
\end{dependency}    
\end{center}
\noindent where \texttt{xsn} is a noun derivational affix.

\subsection{Non-core dependents: nominals}

While \texttt{obl} (oblique nominal) represents an adjunct, in current Korean Universal Dependencies, any arguments with \texttt{jkb} (adverbial postposition) are annotated as \texttt{obl} regardless of whether they are complements of the predicate or not.

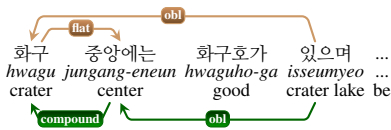
\begin{center}
\begin{dependency}
{
{\scriptsize
   \begin{deptext}
화구\& 중앙에는\& 화구호가 \& 있으며 \&...\\
\textit{hwagu}\& \textit{jungang-eneun} \& \textit{hwaguho-ga} \& \textit{isseumyeo} \&  ... \\
crater\& center \& good \& crater lake \& be\\
   \end{deptext}
    \depedge[theme=copper,edge unit distance=1ex]{4}{1}{obl}   
    \depedge[theme=copper,edge unit distance=1.5ex]{1}{2}{flat}   
   \depedge[theme=grassy,edge below,edge unit distance=1ex]{4}{2}{obl}   
   \depedge[theme=grassy,edge below,edge unit distance=2ex]{2}{1}{compound}   
}
}
\end{dependency}        
\end{center}
However, an argument \textit{jutaegsijang-e} (`housing market') is the complement of the predicate \textit{joh}  (`good'). 
The adjective has the following subcategorization frame (defined in the Sejong dictionary), where it can take an argument as \texttt{goal} associated with a body part, a human, or an abstract object.

{\scriptsize
\begin{verbatim}
X=N0-이 Y=N1-에|에게 좋다 (X=N0-i Y=N1-e|ege johda)
"X"="THM": 구체물|추상적대상 
           (concrete object | abstract object)
"Y"="GOL": 신체부위|인간|추상적대상 
           (body part|human|abstract object)
\end{verbatim}
}

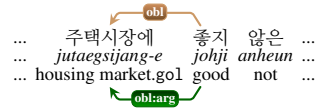
\begin{center}
\begin{dependency}
{
{\scriptsize
   \begin{deptext}
...\& 주택시장에\& 좋지 \& 않은 \&...\\
...\& \textit{jutaegsijang-e} \& \textit{johji} \& \textit{anheun} \&  ... \\
...\& housing market.\texttt{gol} \& good \& not \& ...\\
   \end{deptext}
    \depedge[theme=copper,edge unit distance=1.5ex]{3}{2}{obl}   
    \depedge[theme=grassy,edge below,edge unit distance=1.5ex]{3}{2}{obl:arg}   
}
}
\end{dependency}    
\end{center}
Therefore, we use  the relation \texttt{obl:arg} for oblique arguments and distinguish them from \texttt{obl}.

A \texttt{vocative} relation can be annotated with the vocative marker such as \texttt{jkv} in the GSD corpus or \texttt{jcv} in the Kaist corpus. 
We do not consider \texttt{expl} (expletive) in Korean. 
The \texttt{dislocated} (dislocated elements) relation is annotated for any arguments with a topical marker (\textit{eun/neun}), particularly  in the current Korean Kaist UD.
However, most of currently \texttt{dislocated} annotated nominal arguments should be considered as subjects of a sentence, and we annotate them as \texttt{dislocated:nsubj}.
An example for \texttt{dislocated} is \textit{kokkili-neun} (`elephant.\texttt{top}') $\xleftarrow{\text{\texttt{dislocated}}}$ \textit{gilda} (`long') where \textit{gilda} is a head in a double subject sentence \textit{kokkili-neun ko-ga gilda} (`elephant's nose is long'), and \textit{kokkili-neun} (`elephant.\texttt{top}') actually represents \textit{kokkili-ui} (`elephant.\texttt{gen}').

\subsection{Non-core dependents: function words}

A \texttt{aux} (auxiliary) relation can be associated with an auxiliary predicate that expresses categories such as tense, mood, aspect, voice or evidentiality, which should be distinguished with catenative constructions, as we discussed in $\S$\ref{intro}.

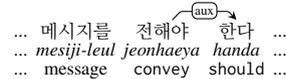
\begin{center}
\begin{dependency}
{
{\scriptsize
   \begin{deptext}
...\& 메시지를\& 전해야 \&한다 \&...\\
...\& \textit{mesiji-leul} \& \textit{jeonhaeya} \& \textit{handa} \&  ... \\
...\& message \&  \texttt{convey} \& \texttt{should} \& ...\\
   \end{deptext}
    \depedge[edge unit distance=1ex]{3}{4}{aux}   
}
}
\end{dependency}    
\end{center}

While copular construction (\texttt{cop}) exists in Korean, it forms words such as \textit{hagsaeng-i} (`student.\texttt{cop}'), where we would not annotate the internal structure of the word. However, \texttt{cop} can be annotated if a copular marker (\textit{-i}) is tokenized from the lexeme due to punctuation marks or symbols.

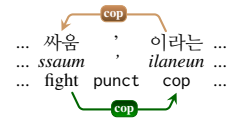
\begin{center}
\begin{dependency}
{
{\scriptsize
   \begin{deptext}
...\& 싸움\& ' \&이라는 \&...\\
...\& \textit{ssaum} \& \textit{'} \& \textit{ilaneun} \&  ... \\
...\& fight \&  \texttt{punct} \& \texttt{cop} \& ...\\
   \end{deptext}
    \depedge[theme=copper,edge unit distance=1ex]{4}{2}{cop}  \depedge[theme=grassy,edge below,edge unit distance=1ex]{2}{4}{cop}   
}
}
\end{dependency}    
\end{center}

A marker (\texttt{mark}) is the functional word marking a clause as subordinate to another clause.

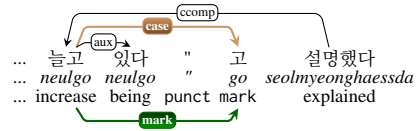
\begin{center}
\begin{dependency}
{
{\scriptsize
   \begin{deptext}
...\& 늘고\& 있다\& " \&고 \&설명했다\\
...\& \textit{neulgo}\& \textit{neulgo} \& \textit{"} \& \textit{go} \&  \textit{seolmyeonghaessda} \\
...\& increase\& being \&  \texttt{punct} \& \texttt{mark} \& explained\\
   \end{deptext}
    \depedge[theme=copper,edge unit distance=.8ex]{2}{5}{case}   
\depedge[edge unit distance=1ex]{6}{2}{ccomp}       
\depedge[edge unit distance=.5ex]{2}{3}{aux}   
    \depedge[theme=grassy,edge below,edge unit distance=.5ex]{2}{5}{mark} 
}
}
\end{dependency}    
\end{center}

\subsection{MWE}

A \texttt{fixed} (fixed grammaticized expressions) relation is used for a certain fixed grammaticized expression, such as \textit{ppun~anila} (`in addition to').

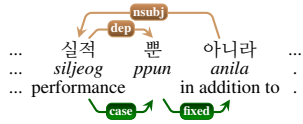
\begin{center}
\begin{dependency}
{
{\scriptsize
   \begin{deptext}
...\& 실적\& 뿐 \&아니라 \&...\\
...\& \textit{siljeog} \& \textit{ppun} \& \textit{anila} \&  . \\
...\& performance \& {} \& in addition to \& .\\
   \end{deptext}
\depedge[theme=copper,edge unit distance=1.5ex]{4}{2}{nsubj}   
\depedge[theme=copper,edge unit distance=1ex]{2}{3}{dep}   
\depedge[theme=grassy,edge below,edge unit distance=1.5ex]{2}{3}{case}
\depedge[theme=grassy,edge below,edge unit distance=1.5ex]{3}{4}{fixed}
}
}
\end{dependency}    
\end{center}

While a \texttt{flat} relation is typically used for headless semi-fixed MWEs like proper names and dates, where it attaches to the first word, such as \textit{Hillary} $\xrightarrow{\text{\texttt{flat}}}$ \textit{Clinton}, we annotate \texttt{flat} to attach to the last word for Korean.

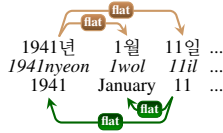
\begin{center}
\begin{dependency}
{
{\scriptsize
   \begin{deptext}
1941년 \& 1월\& 11일 \&...\\
\textit{1941nyeon} \& \textit{1wol} \& \textit{11il} \& ... \\
1941 \& January \& {11} \& ...\\
   \end{deptext}
\depedge[theme=copper,edge unit distance=1.5ex]{1}{2}{flat}   
\depedge[theme=copper,edge unit distance=1.5ex]{1}{3}{flat}   
\depedge[theme=grassy,edge below,edge unit distance=1.5ex]{3}{1}{flat}   
\depedge[theme=grassy,edge below,edge unit distance=1.5ex]{3}{2}{flat}
}
}
\end{dependency}    
\end{center}

The \texttt{compound} relation is primarily used for noun compounds, such as  \textit{guseong wonso} (`constituent element'), but it can be employed for various types of compounding. 

\begin{center}
\begin{dependency}
{
{\scriptsize
   \begin{deptext}
... \& 항성의\& 구성\& 원소에 \& 따라 \& ... \\
...\& \textit{hangseong-ui}\& \textit{guseong}\& \textit{wonso-e} \& \textit{ttala}\& ... \\
...\& star.\texttt{gen} \& constituent\& element \& depending on \& ... \\
   \end{deptext}
\depedge[theme=copper,edge unit distance=2ex]{3}{2}{nmod:poss}   
\depedge[theme=copper,edge unit distance=2ex]{5}{3}{obl}   
\depedge[theme=copper,edge unit distance=2ex]{3}{4}{flat}   
\depedge[theme=grassy,edge below,edge unit distance=2ex]{4}{5}{fixed}
\depedge[theme=grassy,edge below,edge unit distance=2ex]{4}{2}{nmod:poss}
\depedge[theme=grassy,edge below,edge unit distance=2ex]{4}{3}{compound}
}
}
\end{dependency}    
\end{center}



\section{Discussion}

\subsection{Annotating new sentences}
We randomly selected 200 sentences from the Sejong project and manually annotated them using the revised UD guidelines to validate the revisions. 
We use sentence examples from the Sejong verb dictionary, with its frame information where it describes in detail their argument structures alongside their postposition markers. Following sentences are annotated examples that utilize frame information from the Sejong verb dictionary.

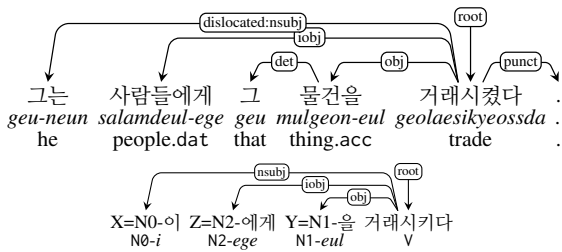
\begin{center}
{\scriptsize
\resizebox{.48\textwidth}{!}{
\begin{dependency}
\begin{deptext}
그는 \& 사람들에게 \& 그 \& 물건을 \& 거래시켰다 \& . \\
\textit{geu-neun} \& \textit{salamdeul-ege} \& \textit{geu} \& \textit{mulgeon-eul} \& \textit{geolaesikyeossda} \& \textit{.} \\
he \&  people.\texttt{dat} \& that \& thing.\texttt{acc} \& trade \&. \\
\end{deptext}
\depedge[edge unit distance=1.5ex]{5}{2}{iobj}
\depedge[edge unit distance=2ex]{4}{3}{det}
\depedge[edge unit distance=2ex]{5}{4}{obj}
\deproot[edge unit distance=2ex]{5}{root}
\depedge[edge unit distance=2ex]{5}{6}{punct}
\depedge[edge unit distance=1.5ex]{5}{1}{dislocated:nsubj}
\end{dependency}}
\begin{dependency}
\begin{deptext}
X=N0-이 \& Z=N2-에게\& Y=N1-을\& 거래시키다 \\
\texttt{N0}-\textit{i} \& \texttt{N2}-\textit{ege} \& \texttt{N1}-\textit{eul} \& \texttt{V} \\
\end{deptext}
\depedge[edge unit distance=1.5ex]{4}{1}{nsubj}
\depedge[edge unit distance=1.5ex]{4}{2}{iobj}
\depedge[edge unit distance=1.5ex]{4}{3}{obj}
\deproot[edge unit distance=1.5ex]{4}{root}
\end{dependency}
}
\end{center}
where \textit{geolaesikida} (`trade') has three arguments: \texttt{N0}-\textit{i} (\texttt{nsubj}), \texttt{N0}-\textit{ege} (\texttt{iobj}) and \texttt{N1}-\textit{e} (\texttt{obj}).

The current UD relations define only \texttt{nsubj}, \texttt{obj} and \texttt{iobj} as nominal core arguments. 
As we discussed previously, we use \texttt{obl:arg} if the predicate takes the adverbial phrase as an argument, as defined in the Sejong dictionary, to distinguish it from the adjunct.

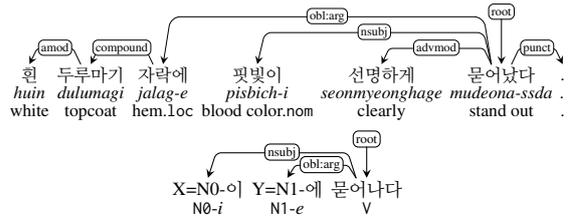
\begin{center}
{\scriptsize
\resizebox{.48\textwidth}{!}{
\begin{dependency}
\begin{deptext}
흰 \& 두루마기 \& 자락에 \& 핏빛이 \& 선명하게 \& 묻어났다 \& . \\
\textit{huin} \& \textit{dulumagi} \& \textit{jalag-e} \& \textit{pisbich-i} \& \textit{seonmyeonghage} \& \textit{mudeona-ssda} \& \textit{.} \\
white \& topcoat \& hem.\texttt{loc} \& blood color.\texttt{nom} \& clearly \& stand out \& .\\
\end{deptext}
\depedge[edge unit distance=2ex]{2}{1}{amod}
\depedge[edge unit distance=2ex]{3}{2}{compound}
\depedge[edge unit distance=2ex]{6}{3}{obl:arg}
\depedge[edge unit distance=2ex]{6}{4}{nsubj}
\depedge[edge unit distance=2ex]{6}{5}{advmod}
\deproot[edge unit distance=2ex]{6}{root}
\depedge[edge unit distance=2ex]{6}{7}{punct}
\end{dependency}
}
\begin{dependency}
\begin{deptext}
X=N0-이 \& Y=N1-에 \& 묻어나다\\
\texttt{N0}-\textit{i} \& \texttt{N1}-\textit{e} \& \texttt{V} \\
\end{deptext}
\depedge[edge unit distance=1.5ex]{3}{2}{obl:arg}
\depedge[edge unit distance=1.5ex]{3}{1}{nsubj}
\deproot[edge unit distance=1.5ex]{3}{root}
\end{dependency}
}
\end{center}
where \textit{mudeonada} (`stand out') has two arguments: \texttt{N0}-\textit{i} (\texttt{nsubj}) and \texttt{N1}-\textit{e} (\texttt{obl:arg}).

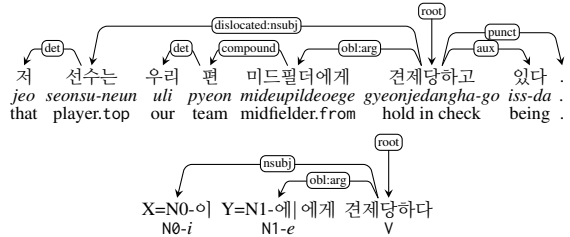
\begin{center}
{\scriptsize
\resizebox{.48\textwidth}{!}{
\begin{dependency}
\begin{deptext}
저 \& 선수는 \& 우리 \& 편 \& 미드필더에게 \& 견제당하고 \& 있다 \& . \\
\textit{jeo} \& \textit{seonsu-neun} \& \textit{uli} \& \textit{pyeon} \& \textit{mideupildeoege} \& \textit{gyeonjedangha-go} \& \textit{iss-da} \& \textit{.} \\
{that} \& {player.\texttt{top}} \& {our} \& {team} \& {midfielder.\texttt{from}} \& {hold in check} \& {being} \& {.} \\
\end{deptext}
\depedge[edge unit distance=2ex]{2}{1}{det}
\depedge[edge unit distance=2ex]{4}{3}{det}
\depedge[edge unit distance=2ex]{5}{4}{compound}
\depedge[edge unit distance=1.2ex]{6}{2}{dislocated:nsubj}
\depedge[edge unit distance=2ex]{6}{5}{obl:arg}
\deproot[edge unit distance=2ex]{6}{root}
\depedge[edge unit distance=2ex]{6}{7}{aux}
\depedge[edge unit distance=2ex]{6}{8}{punct}
\end{dependency}
}
\begin{dependency}
\begin{deptext}
X=N0-이 \& Y=N1-에$|$에게 \& 견제당하다 \\
\texttt{N0}-\textit{i} \& \texttt{N1}-\textit{e} \& \texttt{V} \\
\end{deptext}
\depedge[edge unit distance=2ex]{3}{1}{nsubj}
\depedge[edge unit distance=2ex]{3}{2}{obl:arg}
\deproot[edge unit distance=2ex]{3}{root}
\end{dependency}
}
\end{center}
where \textit{gyeonjedangha} (`hold in check') has two arguments: \texttt{N0}-\textit{i} (\texttt{nsubj}) and \texttt{N1}-\textit{e$|$ege} (\texttt{obl:arg}).

\subsection{Comparing with Sejong-style dependency structure}

The dependency structure for Korean has been established based on the Sejong constituency treebank, with previous work converting it into a dependency treebank \citep{oh-cha:2013,park-EtAl:2013:IWPT}. Accordingly, tokenization and dependency relation notation are derived from the Sejong constituent treebank. In this framework, the \textit{eojeol} (a space unit in Korean) serves as the basic analysis unit, and punctuation marks are not separated from the word. Moreover, phrase labels from the Sejong treebank are used for annotating dependency relations. 
Recent KLUE benchmark \citep{park-etal-2021-klue} also follows the Sejong-style dependency structure.

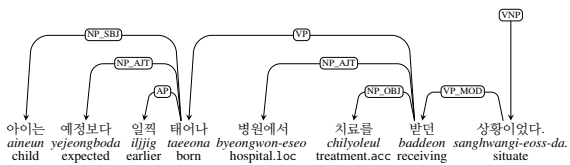
\begin{center}
{\scriptsize
\resizebox{.48\textwidth}{!}{
\begin{dependency}
\begin{deptext}
아이는 \&  예정보다 \&  일찍 \&  태어나 \&  병원에서 \&  치료를 \&  받던 \&  상황이었다. \\
\textit{aineun} \& \textit{yejeongboda} \& \textit{iljjig} \& \textit{taeeona} \& \textit{byeongwon-eseo } \& \textit{chilyoleul} \& \textit{baddeon}\& \textit{sanghwangi-eoss-da.} \\
{child} \& {expected} \& {earlier} \& {born} \& {hospital.\texttt{loc}} \& {treatment.\texttt{acc}} \& {receiving} \& {situate} \\
\end{deptext}
\depedge{4}{1}{NP\_SBJ}
\depedge{4}{2}{NP\_AJT}
\depedge{4}{3}{AP}
\depedge{7}{4}{VP}
\depedge{7}{5}{NP\_AJT}
\depedge{7}{6}{NP\_OBJ}
\depedge{8}{7}{VP\_MOD}
\deproot{8}{VNP}
\end{dependency}
}}
\end{center}
where \texttt{NP\_SBJ}, \texttt{NP\_AJT} and \texttt{NP\_OBJ} represent a nominal subject, an adjunct, and an object, respectively. 
Instead of \texttt{root}, it indicates the property of the root, such as \texttt{VNP} (copular), and the last word \textit{sanghwangi-eoss-da.} (`situate') includes the punctuation mark.
Most importantly, the dependency structure consistently adheres to a right-to-left pattern, affirming that Korean is an end-focus language. 

\section{Conclusion}

{We are planning to integrate full and detailed guidelines with Korean sentence examples into the Universal Dependency Relations pages for Korean, which are currently unavailable.\footnote{\url{https://universaldependencies.org/u/dep/}}}
{We are also in the process of aligning the UD-style and Sejong-style dependency treebanks as part of our efforts to establish \textbf{a linguistic consensus model} for the analysis of the Korean language in collaboration with the KLUE benchmark consortium and the National Institute of Korean Language.}



\begin{thebibliography}{19}
\expandafter\ifx\csname natexlab\endcsname\relax\def\natexlab#1{#1}\fi

\bibitem[{Chen et~al.(2022)Chen, Jo, Yao, Lim, Silfverberg, Tyers, and Park}]{chen-etal-2022-yet}
Yige Chen, Eunkyul~Leah Jo, Yundong Yao, KyungTae Lim, Miikka Silfverberg, Francis~M. Tyers, and Jungyeul Park. 2022.
\newblock \href {https://aclanthology.org/2022.coling-1.482} {{Yet Another Format of Universal Dependencies for Korean}}.
\newblock In \emph{Proceedings of the 29th International Conference on Computational Linguistics}, pages 5432--5437, Gyeongju, Republic of Korea. International Committee on Computational Linguistics.

\bibitem[{Choi and Palmer(2011)}]{choi-palmer:2011:SPMRL}
Jinho~D. Choi and Martha Palmer. 2011.
\newblock \href {http://www.aclweb.org/anthology/W11-3801} {{Statistical Dependency Parsing in Korean: From Corpus Generation To Automatic Parsing}}.
\newblock In \emph{Proceedings of the Second Workshop on Statistical Parsing of Morphologically Rich Languages}, pages 1--11, Dublin, Ireland. Association for Computational Linguistics.

\bibitem[{Choi et~al.(1994)Choi, Han, Han, and Kwon}]{choi-EtAl:1994}
Key-Sun Choi, Young~S. Han, Young~G. Han, and Oh~W. Kwon. 1994.
\newblock {KAIST Tree Bank Project for Korean: Present and Future Development}.
\newblock In \emph{Proceedings of the International Workshop on Sharable Natural Language Resources}, pages 7--14, Nara Institute of Science and Technology. Nara Institute of Science and Technology.

\bibitem[{Chun et~al.(2018)Chun, Han, Hwang, and Choi}]{chun-EtAl:2018:LREC}
Jayeol Chun, Na-Rae Han, Jena~D. Hwang, and Jinho~D. Choi. 2018.
\newblock {Building Universal Dependency Treebanks in Korean}.
\newblock In \emph{Proceedings of the Eleventh International Conference on Language Resources and Evaluation (LREC 2018)}, Miyazaki, Japan. European Language Resources Association (ELRA).

\bibitem[{de~Marneffe et~al.(2014)de~Marneffe, Dozat, Silveira, Haverinen, Ginter, Nivre, and Manning}]{deMarneffe-EtAl:2014:LREC}
Marie-Catherine de~Marneffe, Timothy Dozat, Natalia Silveira, Katri Haverinen, Filip Ginter, Joakim Nivre, and Christopher~D. Manning. 2014.
\newblock \href {http://www.lrec-conf.org/proceedings/lrec2014/pdf/1062_Paper.pdf} {{Universal Stanford dependencies: A cross-linguistic typology}}.
\newblock In \emph{Proceedings of the Ninth International Conference on Language Resources and Evaluation (LREC'14)}, pages 4585--4592, Reykjavik, Iceland. European Language Resources Association (ELRA).

\bibitem[{Han et~al.(2002)Han, Han, Ko, Palmer, and Yi}]{han-EtAl:2002}
Chung-Hye Han, Na-Rae Han, Eon-Suk Ko, Martha Palmer, and Heejong Yi. 2002.
\newblock {Penn Korean Treebank: Development and Evaluation}.
\newblock In \emph{Proceedings of the 16th Pacific Asia Conference on Language, Information and Computation}, pages 69--78, Jeju, Korea. Pacific Asia Conference on Language, Information and Computation.

\bibitem[{Jo et~al.(2023)Jo, Kim, Wu, Lim, Park, and Park}]{jo-etal-2023-k}
Eunkyul Jo, Kyuwon Kim, Xihan Wu, KyungTae Lim, Jungyeul Park, and Chulwoo Park. 2023.
\newblock \href {https://aclanthology.org/2023.findings-acl.414} {{K-UniMorph: Korean Universal Morphology and its Feature Schema}}.
\newblock In \emph{Findings of the Association for Computational Linguistics: ACL 2023}, pages 6613--6623, Toronto, Canada. Association for Computational Linguistics.

\bibitem[{Kanayama et~al.(2014)Kanayama, Park, Tsuboi, and Yi}]{kanayama-EtAl:2014:LT4CloseLang}
Hiroshi Kanayama, Youngja Park, Yuta Tsuboi, and Dongmook Yi. 2014.
\newblock \href {https://doi.org/10.3115/v1/W14-4202} {{Learning from a Neighbor: Adapting a Japanese Parser for Korean Through Feature Transfer Learning}}.
\newblock In \emph{Proceedings of the EMNLP'2014 Workshop on Language Technology for Closely Related Languages and Language Variants}, pages 2--12, Doha, Qatar. Association for Computational Linguistics.

\bibitem[{McDonald et~al.(2013)McDonald, Nivre, Quirmbach-Brundage, Goldberg, Das, Ganchev, Hall, Petrov, Zhang, T{\"{a}}ckstr{\"{o}}m, Bedini, Bertomeu~Castell{\'{o}}, and Lee}]{mcdonald-etal-2013-universal}
Ryan McDonald, Joakim Nivre, Yvonne Quirmbach-Brundage, Yoav Goldberg, Dipanjan Das, Kuzman Ganchev, Keith Hall, Slav Petrov, Hao Zhang, Oscar T{\"{a}}ckstr{\"{o}}m, Claudia Bedini, Núria Bertomeu~Castell{\'{o}}, and Jungmee Lee. 2013.
\newblock \href {https://aclanthology.org/P13-2017} {{Universal Dependency Annotation for Multilingual Parsing}}.
\newblock In \emph{Proceedings of the 51st Annual Meeting of the Association for Computational Linguistics (Volume 2: Short Papers)}, pages 92--97, Sofia, Bulgaria. Association for Computational Linguistics.

\bibitem[{Nivre et~al.(2016)Nivre, de~Marneffe, Ginter, Goldberg, Hajic, Manning, McDonald, Petrov, Pyysalo, Silveira, Tsarfaty, and Zeman}]{nivre-EtAl:2016:LREC}
Joakim Nivre, Marie-Catherine de~Marneffe, Filip Ginter, Yoav Goldberg, Jan Hajic, Christopher~D. Manning, Ryan McDonald, Slav Petrov, Sampo Pyysalo, Natalia Silveira, Reut Tsarfaty, and Daniel Zeman. 2016.
\newblock \href {https://www.aclweb.org/anthology/L16-1262%0A} {{Universal Dependencies v1: A Multilingual Treebank Collection}}.
\newblock In \emph{Proceedings of the Tenth International Conference on Language Resources and Evaluation (LREC 2016)}, page 1659–1666, Portoro{\v{z}}, Slovenia. European Language Resources Association (ELRA).

\bibitem[{Nivre et~al.(2020)Nivre, de~Marneffe, Ginter, Haji{\v{c}}, Manning, Pyysalo, Schuster, Tyers, and Zeman}]{nivre-EtAl:2020:LREC}
Joakim Nivre, Marie-Catherine de~Marneffe, Filip Ginter, Jan Haji{\v{c}}, Christopher~D. Manning, Sampo Pyysalo, Sebastian Schuster, Francis Tyers, and Daniel Zeman. 2020.
\newblock \href {https://www.aclweb.org/anthology/2020.lrec-1.497} {{Universal Dependencies v2: An Evergrowing Multilingual Treebank Collection}}.
\newblock In \emph{Proceedings of the 12th Language Resources and Evaluation Conference}, pages 4034--4043, Marseille, France. European Language Resources Association.

\bibitem[{Noh et~al.(2018)Noh, Han, Oh, and Kim}]{noh-etal-2018-enhancing}
Youngbin Noh, Jiyoon Han, Tae~Hwan Oh, and Hansaem Kim. 2018.
\newblock \href {https://doi.org/10.18653/v1/W18-6013} {{Enhancing Universal Dependencies for Korean}}.
\newblock In \emph{Proceedings of the Second Workshop on Universal Dependencies (UDW 2018)}, pages 108--116, Brussels, Belgium. Association for Computational Linguistics.

\bibitem[{Oh and Cha(2013)}]{oh-cha:2013}
Jin-Young Oh and Jeong-Won Cha. 2013.
\newblock {Korean Dependency Parsing using Key Eojoel}.
\newblock \emph{Journal of KIISE:Software and Applications}, 40(10):600--608.

\bibitem[{Park et~al.(2013)Park, Kawahara, Kurohashi, and Choi}]{park-EtAl:2013:IWPT}
Jungyeul Park, Daisuke Kawahara, Sadao Kurohashi, and Key-Sun Choi. 2013.
\newblock \href {https://www.aclweb.org/anthology/W13-5714} {{Towards Fully Lexicalized Dependency Parsing for Korean}}.
\newblock In \emph{Proceedings of the 13th International Conference on Parsing Technologies (IWPT 2013)}, pages 120--126, Nara, Japan. Assocation for Computational Linguistics.

\bibitem[{Park et~al.(2021)Park, Moon, Kim, Cho, Han, Park, Song, Kim, Song, Oh, Lee, Oh, Lyu, Jeong, Lee, Seo, Lee, Kim, Lee, Jang, Do, Kim, Lim, Lee, Park, Shin, Kim, Park, Oh, Ha, and Cho}]{park-etal-2021-klue}
Sungjoon Park, Jihyung Moon, Sungdong Kim, Won~Ik Cho, Ji~Yoon Han, Jangwon Park, Chisung Song, Junseong Kim, Youngsook Song, Taehwan Oh, Joohong Lee, Juhyun Oh, Sungwon Lyu, Younghoon Jeong, Inkwon Lee, Sangwoo Seo, Dongjun Lee, Hyunwoo Kim, Myeonghwa Lee, Seongbo Jang, Seungwon Do, Sunkyoung Kim, Kyungtae Lim, Jongwon Lee, Kyumin Park, Jamin Shin, Seonghyun Kim, Lucy Park, Alice Oh, Jung-Woo Ha, and Kyunghyun Cho. 2021.
\newblock \href {https://datasets-benchmarks-proceedings.neurips.cc/paper_files/paper/2021/file/98dce83da57b0395e163467c9dae521b-Paper-round2.pdf} {{KLUE: Korean Language Understanding Evaluation}}.
\newblock In \emph{Proceedings of the Neural Information Processing Systems Track on Datasets and Benchmarks}, volume~1, pages 1--25. Curran.

\bibitem[{Seddah et~al.(2014)Seddah, K{\"{u}}bler, and Tsarfaty}]{seddah-kubler-tsarfaty:2014}
Djamé Seddah, Sandra K{\"{u}}bler, and Reut Tsarfaty. 2014.
\newblock \href {https://www.aclweb.org/anthology/W14-6111} {{Introducing the SPMRL 2014 Shared Task on Parsing Morphologically-rich Languages}}.
\newblock In \emph{Proceedings of the First Joint Workshop on Statistical Parsing of Morphologically Rich Languages and Syntactic Analysis of Non-Canonical Languages}, pages 103--109, Dublin, Ireland. Dublin City University.

\bibitem[{Seddah et~al.(2013)Seddah, Tsarfaty, K{\"{u}}bler, Candito, Choi, Farkas, Foster, Goenaga, Gojenola~Galletebeitia, Goldberg, Green, Habash, Kuhlmann, Maier, Nivre, Przepi{\'{o}}rkowski, Roth, Seeker, Versley, Vincze, Woli{\'{n}}ski, Wr{\'{o}}blewska, and de~la Clergerie}]{seddah-EtAl:2013:SPMRL}
Djamé Seddah, Reut Tsarfaty, Sandra K{\"{u}}bler, Marie Candito, Jinho~D. Choi, Richárd Farkas, Jennifer Foster, Iakes Goenaga, Koldo Gojenola~Galletebeitia, Yoav Goldberg, Spence Green, Nizar Habash, Marco Kuhlmann, Wolfgang Maier, Joakim Nivre, Adam Przepi{\'{o}}rkowski, Ryan Roth, Wolfgang Seeker, Yannick Versley, Veronika Vincze, Marcin Woli{\'{n}}ski, Alina Wr{\'{o}}blewska, and Eric~Villemonte de~la Clergerie. 2013.
\newblock \href {http://www.aclweb.org/anthology/W13-4917} {{Overview of the SPMRL 2013 Shared Task: A Cross-Framework Evaluation of Parsing Morphologically Rich Languages}}.
\newblock In \emph{Proceedings of the Fourth Workshop on Statistical Parsing of Morphologically-Rich Languages}, pages 146--182, Seattle, Washington, USA. Association for Computational Linguistics.

\bibitem[{Zeman et~al.(2018)Zeman, Haji{\v{c}}, Popel, Potthast, Straka, Ginter, Nivre, and Petrov}]{zeman-etal-2018-conll}
Daniel Zeman, Jan Haji{\v{c}}, Martin Popel, Martin Potthast, Milan Straka, Filip Ginter, Joakim Nivre, and Slav Petrov. 2018.
\newblock \href {https://doi.org/10.18653/v1/K18-2001} {{CoNLL 2018 Shared Task: Multilingual Parsing from Raw Text to Universal Dependencies}}.
\newblock In \emph{Proceedings of the CoNLL 2018 Shared Task: Multilingual Parsing from Raw Text to Universal Dependencies}, pages 1--21, Brussels, Belgium. Association for Computational Linguistics.

\bibitem[{Zeman et~al.(2017)Zeman, Popel, Straka, Haji{\v{c}}, Nivre, Ginter, Luotolahti, Pyysalo, Petrov, Potthast, Tyers, Badmaeva, Gokirmak, Nedoluzhko, Cinkov{\'{a}}, Haji{\v{c}}~Jr., Hlav{\'{a}}{\v{c}}ov{\'{a}}, Kettnerov{\'{a}}, Ure{\v{s}}ov{\'{a}}, Kanerva, Ojala, Missil{\"{a}}, Manning, Schuster, Reddy, Taji, Habash, Leung, Marneffe, Sanguinetti, Simi, Kanayama, Paiva, Droganova, Alonso, {\c{C}}{\"{o}}ltekin, Sulubacak, Uszkoreit, Macketanz, Burchardt, Harris, Marheinecke, Rehm, Kayadelen, Attia, Elkahky, Yu, Pitler, Lertpradit, Mandl, Kirchner, Alcalde, Strnadov{\'{a}}, Banerjee, Manurung, Stella, Shimada, Kwak, Mendon{\c{c}}a, Lando, Nitisaroj, and Li}]{zeman-etal-2017-conll}
Daniel Zeman, Martin Popel, Milan Straka, Jan Haji{\v{c}}, Joakim Nivre, Filip Ginter, Juhani Luotolahti, Sampo Pyysalo, Slav Petrov, Martin Potthast, Francis Tyers, Elena Badmaeva, Memduh Gokirmak, Anna Nedoluzhko, Silvie Cinkov{\'{a}}, Jan Haji{\v{c}}~Jr., Jaroslava Hlav{\'{a}}{\v{c}}ov{\'{a}}, Václava Kettnerov{\'{a}}, Zdeňka Ure{\v{s}}ov{\'{a}}, Jenna Kanerva, Stina Ojala, Anna Missil{\"{a}}, Christopher~D. Manning, Sebastian Schuster, Siva Reddy, Dima Taji, Nizar Habash, Herman Leung, Marie-Catherine~de Marneffe, Manuela Sanguinetti, Maria Simi, Hiroshi Kanayama, Valeria~de Paiva, Kira Droganova, Héctor~Martínez Alonso, {\c{C}}ağrı {\c{C}}{\"{o}}ltekin, Umut Sulubacak, Hans Uszkoreit, Vivien Macketanz, Aljoscha Burchardt, Kim Harris, Katrin Marheinecke, Georg Rehm, Tolga Kayadelen, Mohammed Attia, Ali Elkahky, Zhuoran Yu, Emily Pitler, Saran Lertpradit, Michael Mandl, Jesse Kirchner, Hector~Fernandez Alcalde, Jana Strnadov{\'{a}}, Esha Banerjee, Ruli Manurung, Antonio Stella, Atsuko Shimada,
  Sookyoung Kwak, Gustavo Mendon{\c{c}}a, Tatiana Lando, Rattima Nitisaroj, and Josie Li. 2017.
\newblock \href {https://doi.org/10.18653/v1/K17-3001} {{CoNLL 2017 Shared Task: Multilingual Parsing from Raw Text to Universal Dependencies}}.
\newblock In \emph{Proceedings of the CoNLL 2017 Shared Task: Multilingual Parsing from Raw Text to Universal Dependencies}, pages 1--19, Vancouver, Canada. Association for Computational Linguistics.

\end{thebibliography}




\end{document}